\documentclass[]{fairmeta}
\usepackage{amssymb}
\usepackage{multirow}
\usepackage{bigdelim}
\usepackage{todonotes}
\usepackage{longtable}
\usepackage{tabularray}
\usepackage{wrapfig}
\usepackage[most]{tcolorbox} 
\usepackage{xcolor}
\usepackage{url}

\usepackage[utf8]{inputenc}
\usepackage{tikz}
\usepackage{amsmath}
\usepackage{caption}
\usepackage{subcaption}
\usepackage[table]{xcolor}

\usepackage{pifont}
\usepackage{makecell}
\usepackage{array} % for table
\usepackage{makecell}
\usepackage{ulem}

\usepackage{siunitx}
\newcommand{\cmark}{\ding{51}}
\newcommand{\xmark}{\ding{55}}

\makeatletter
\newcommand{\onedot}{\@ifnextchar.{}{.}}
\makeatother

\def\eg{\emph{e.g}\onedot} 

\def\ie{\emph{i.e}\onedot}

\usetikzlibrary{shapes.geometric, arrows.meta, positioning, decorations.pathmorphing}

\definecolor{predblue}{RGB}{66,153,225}
\definecolor{gtgreen}{RGB}{72,187,120}
\definecolor{matchgray}{RGB}{160,174,192}
\definecolor{textgray}{RGB}{74,85,104}
\definecolor{titlegray}{RGB}{45,55,72}

\tikzset{
    mydiamond/.style={
        diamond,
        fill=#1,
        draw=#1,
        minimum size=6pt,
        inner sep=0pt
    }
}

\definecolor{prompt}{HTML}{5f84e4}
\definecolor{img}{HTML}{820100}

\setcitestyle{numbers,square}

\title{ARC-Chapter: Structuring Hour-Long Videos into Navigable Chapters and Hierarchical Summaries}

\author[*]{Junfu Pu}
\author[*]{Teng Wang}
\author[\dagger]{Yixiao Ge}
\author[]{Yuying Ge}
\author[]{Chen Li}
\author[]{Ying Shan}

\affiliation[]{ARC Lab, Tencent PCG}

\contribution[*]{Core contributors}
\contribution[\dagger]{Project lead}

\abstract{The proliferation of hour-long videos (\eg, lectures, podcasts, documentaries) has intensified demand for efficient content structuring. However, existing approaches are constrained by small-scale training with annotations that are typical short and coarse, restricting generalization to nuanced transitions in long videos. We introduce ARC-Chapter, the first large-scale video chaptering model trained on over million-level long video chapters, featuring bilingual, temporally grounded, and hierarchical chapter annotations. To achieve this goal, we curated a bilingual English-Chinese chapter dataset via a structured pipeline that unifies ASR transcripts, scene texts, visual captions into multi-level annotations, from short title to long summaries. We demonstrate clear performance improvements with data scaling, both in data volume and label intensity. Moreover, we design a new evaluation metric termed GRACE, which incorporates many-to-one segment overlaps and semantic similarity, better reflecting real-world chaptering flexibility. Extensive experiments demonstrate that ARC-Chapter establishes a new state-of-the-art by a significant margin, outperforming the previous best by 14.0\% in F1 score and 11.3\% in SODA score. Moreover, ARC-Chapter shows excellent transferability, improving the state-of-the-art on downstream tasks like dense video captioning on YouCook2.
}

\date{November 18, 2025}

\metadata[Github]{\url{https://github.com/TencentARC/ARC-Chapter}}

\begin{document}

\maketitle

\begin{figure}
    \centering
    \vspace{0mm}  
    \includegraphics[width=1.0\textwidth]{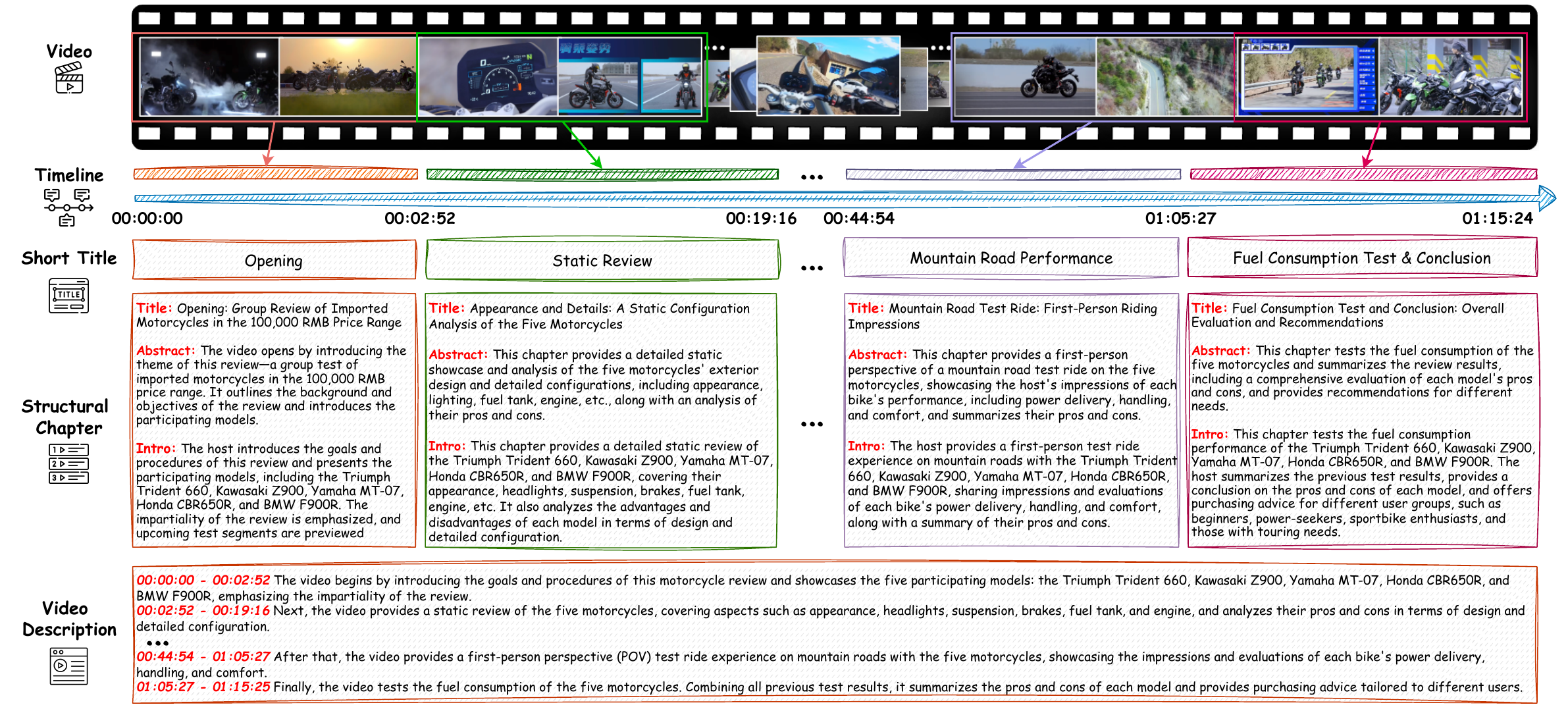}
    \caption{\textbf{An illustration of the capabilities of our video chaptering model.} Given a video, our model is able to generate timestamped chapters with three-level structured output: 1) \textbf{Short Title} - a concise label summarizing each chapter; 2) \textbf{Structural Chapter} - a detailed, structured annotation for each chapter, including a rewritten comprehensive \emph{title}, an \emph{abstract} summarizing the core content, and an \emph{introduction} describing key details and highlights; and 3) \textbf{Timestamp-Aligned Video Description} - fine-grained descriptions aligned with precise temporal boundaries. This hierarchical structure facilitates an efficient and precise understanding of video content.}
  \label{fig:teaser} 
\end{figure}

\section{Introduction}
\label{sec:intro}

The exponential proliferation of long-form video content, including educational lectures, vlogs, live streams, and meeting recordings—poses significant challenges for automatic content understanding. Video chaptering~\cite{ventura2025chapter,vidchapter} has emerged as a promising solution, segmenting videos into navigable and semantically coherent chapters. This enables efficient content retrieval, summarization, and enhanced user interaction, which are critical for managing and consuming large-scale video data.

Despite notable advances in segmenting short videos (usually within five minutes) for tasks such as action segmentation~\cite{ding2023temporal,lea2017temporal,lu2024fact,shen2024progress,wang2020boundary}, temporal event localization~\cite{proposal1,proposal6}, and dense video captioning~\cite{anet,pdvc,yang2023vid2seq}, the structuring of hour-long videos remains a formidable challenge. First, modeling sophisticated semantics across multimodal inputs, including visual and audio streams—over extended temporal horizons requires robust and scalable architectures. Second, the scarcity of large-scale datasets with fine-grained annotations hinders the development and evaluation of effective chaptering models. Third, existing evaluation metrics~\cite{fujita2020soda,anet} often fail to capture the semantic granularity of chapter boundaries, leading to suboptimal matching and similarity scoring between predicted and ground-truth segments~\cite{fujita2020soda}.

In this technical report, we introduce ARC-Chapter, a comprehensive framework designed to address the unique challenges of long-form video structuring. As illustrated in Fig.~\ref{fig:teaser}, ARC-Chapter enables the segmentation of lengthy videos into navigable chapters and generates hierarchical summaries that capture both coarse and fine-grained content structure. Our work makes three primary contributions. First, we advance the scalability of video chaptering by developing the first large-scale model trained on one million long videos, totaling 400,000 hours of content. This dataset is fifty times larger than those used in previous studies~\cite{ventura2025chapter}, allowing our model to generalize across diverse video domains and formats. Second, we propose a semi-automatic annotation pipeline for hierarchical summaries, which leverages easily accessible human-annotated coarse labels. This pipeline integrates automatic speech recognition (ASR) derived transcripts with timestamped visual elements, enabling a holistic and multimodal understanding of video content. Third, we introduce GRACE, a novel granularity-robust evaluation metric designed to address the semantic misalignment issues prevalent in existing chaptering benchmarks. GRACE provides a more accurate assessment of chapter boundary quality by accounting for varying levels of semantic granularity.

Our extensive experiments demonstrate the effectiveness of ARC-Chapter, which establishes a new state-of-the-art on both Chinese and English long-form video chaptering benchmarks. Specifically, ARC-Chapter substantially outperforms previous methods on the VidChapters-7M test sets (e.g., CIDEr: 100.9$\rightarrow$186.6; F1: 45.3$\rightarrow$59.3; SODA: 19.3$\rightarrow$30.6). We validate the importance of multimodality, showing that our full model surpasses video-only and audio-only variants by 7.7 and 5.3 points on SODA, respectively. Furthermore, pretraining on our large-scale dataset significantly enhances transferability, evidenced by notable performance gains on downstream tasks like YouCook2 and ActivityNet Captions. Crucially, our work is the first to identify a clear scaling law in video chaptering: model performance consistently improves with increased training data and label density. This finding refutes previous observations that performance saturates on smaller datasets ($\sim$20k samples)~\cite{ventura2025chapter} and suggests a promising direction for future research.

The remainder of this report is structured as follows: Section~\ref{sec:related_works} reviews related works; Section~\ref{sec:dataset} describes the dataset and annotation pipeline; Section~\ref{sec:method} details our methodology and model architecture; Section~\ref{sec:experiment} presents experimental results and analysis; Section \ref{sec:conclusion} concludes.

\section{Related Works}
\label{sec:related_works}

\paragraph{\textbf{Global Video Understanding. }}
Early video understanding~\cite{abouelenin2025phi,comanici2025gemini,guo2025seed1,li2023otter,lu2024deepseekvl,shu2023audio,wang2025skywork,wu2023nextgpt,yan2024visa,zhang2025videollama3,zhang2025towards,zhang2024llava,zhu2025internvl3} research primarily targeted global comprehension tasks, such as video question answering, video captioning, and video classification. These methods treat entire videos as holistic units, extracting global representations to predict semantic labels or generate summaries. While effective for short videos, they often fail to capture complex temporal dynamics and hierarchical structures of long-form content~\cite{li2024llama,ren2024timechat}.

\paragraph{\textbf{Temporal Segmentation for Short Videos.}}
To address the limitations of global approaches, recent works~\cite{guo2024trace, guo2025vtg, huang2024vtimellm, qian2024momentor, ren2024timechat, dibs, yang2025timeexpert,zhang2025thinking, zhou2024streaming} have shifted towards modeling the temporal structure of videos. Datasets like ActivityNet Captions~\cite{anet}, Charades-STA~\cite{gao2017charade-sta}, YouCook2~\cite{zhou2018youcook2} and Breakfast~\cite{breakfast} provide timestamped event annotations, enabling tasks such as temporal event localization, action segmentation, and dense video captioning. These approaches move beyond global representations to identify and describe fine-grained events and local temporal dependencies.
However, most temporally-structured datasets~\cite{liu2025videomind, zeng2024timesuite} are limited to short clips, typically under several minutes, and thus do not capture the challenges of ultra-long videos found in lectures, podcasts, or livestreams. 
The lack of large-scale, long-duration datasets with fine-grained temporal annotations remains a major bottleneck.

\paragraph{\textbf{Long-Form Video Structuring.}}
A few efforts~\cite{ventura2025chapter,yang2023vidchapters} have explored the structuring of hour-long videos. 
The VidChapters-7M dataset~\cite{yang2023vidchapters} provides a large-scale benchmark for video chaptering, with millions of videos and annotated chapter boundaries, better reflecting real-world scenarios such as vlogs, podcasts, and meetings where long-term temporal reasoning is essential.

Despite these advances, significant challenges remain. Existing chaptering models often rely on limited modalities, such as automatic speech recognition, are trained on small-scale datasets, and produce coarse, uninformative descriptions, which limits their scalability across diverse video domains. To address these issues, we propose a scalable, multimodal framework for long-form video chaptering, supported by a large-scale dataset with detailed chapter descriptions.

\section{Data Collection and Annotation}
\label{sec:dataset}

A significant challenge in developing strong video chaptering models is the scarcity of publicly available datasets with detailed, multi-level annotations.
Existing datasets typically provide only sparse labels, such as video-level categories for video classification or coarse temporal segments with brief titles such as VidChapters-7M. To address this limitation and to facilitate research on hierarchical video chaptering and summarization, we introduce a new, richly annotated video chaptering dataset. This section details our data curation and annotation pipeline.

\begin{figure}
    \centering
    \vspace{0mm}  
    \includegraphics[width=0.99\textwidth]{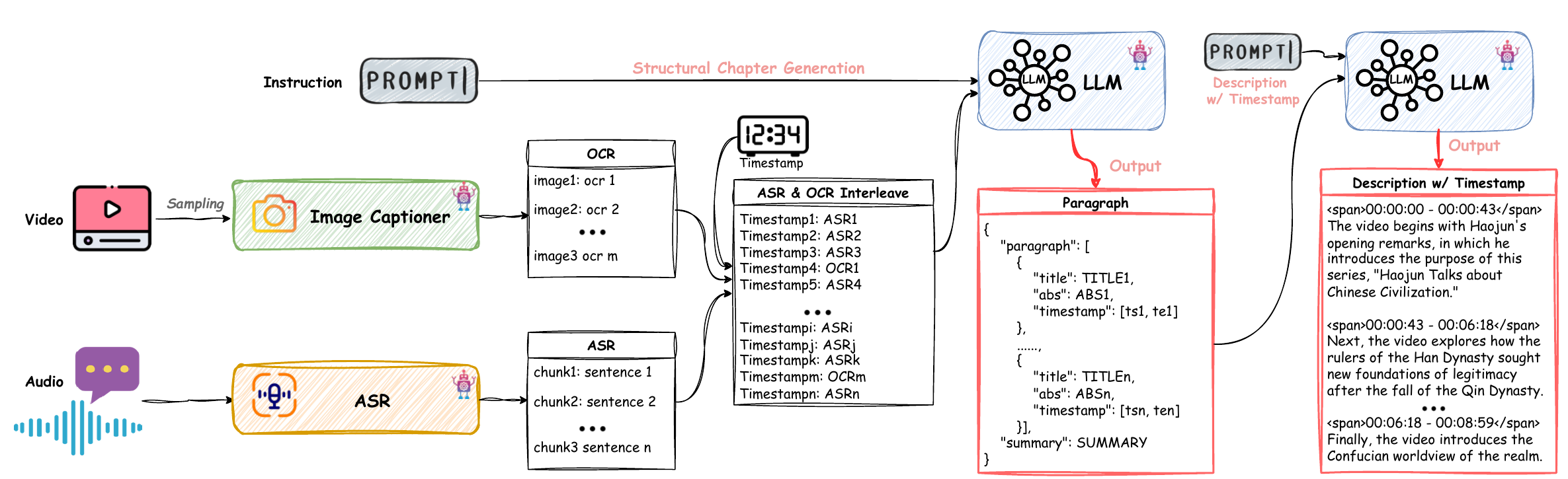}
    \caption{Overview of our automatic video annotation pipeline for hierarchical chaptering and summarization. We extract visual captions (OCR included) from sampled video frames and ASR transcripts from audio. These outputs are temporally aligned and interleaved into a unified multimodal transcript. This transcript, together with original chapter markers, is processed by an LLM to produce structured chapters and timestamp-aligned video descriptions.}
  \label{fig:datapipeline} 
\end{figure}

\subsection{Data Curation}

One of the key contributions of our work is the introduction of a new large-scale dataset, named VidAtlas, which is designed for the task of hierarchical video chaptering and summarization.
Our primary goal is to construct a dataset that not only provides accurate chapter boundaries but also offers dense, multi-granularity textual descriptions for both individual chapters and the entire video.

\paragraph{\textbf{Data Sourcing.}} We begin by sourcing videos from the video platform. The primary selection criterion is the presence of author-provided chapter markers. These markers, which include the start/end timestamps and a short title for each chapter, are manually defined by the video uploader. This approach provides us with a highly accurate human-verified ground truth for the temporal segmentation of videos, which is a significant foundation for our subsequent annotation efforts.
The collected videos, which are long, well-structured, and information-dense, are ideal candidates for video chaptering.

\paragraph{\textbf{Filtering and Refinement.}} Starting with this initial collection, we apply several filtering criteria to guarantee the quality and diversity of our dataset for video understanding and chaptering. First, we retain videos whose durations lie between 2 minutes and 3 hours. This range excludes trivial short clips, which are unnecessary for chaptering, as well as overly long videos, which are often unstructured (e.g., live streams) and difficult to process due to the context-length limitations of our model. Second, we curate videos across a wide range of domains, including educational lectures, DIY tutorials, reviews \& unboxings, interviews \& podcasts, webinars \& presentations, gaming \& music albums, fitness \& cooking and documentaries. This wide distribution of domains ensures that the dataset is not biased towards any specific genre and supports the development of more generalizable models.

\begin{figure}[htbp]
    \centering % 将整个图例（包含所有子图）居中

    % --- 第一个子图 ---
    \begin{subfigure}[b]{0.3\textwidth}
        \centering
        \includegraphics[width=\linewidth]{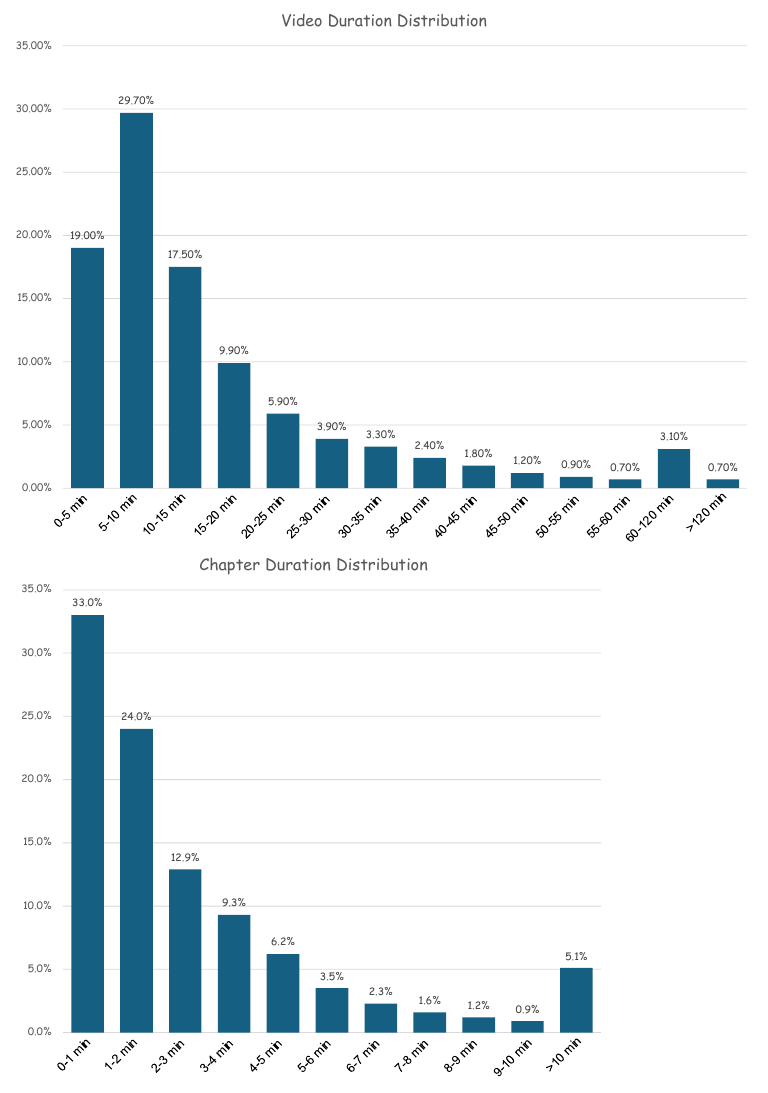}
        \caption{Duration distribution}
        \label{fig:dataset_dis}
    \end{subfigure}
    \hfill % 这个命令在两个子图之间创建弹性空白，将它们推向两边
    \begin{subfigure}[b]{0.68\textwidth}
        \centering
        \includegraphics[width=\linewidth]{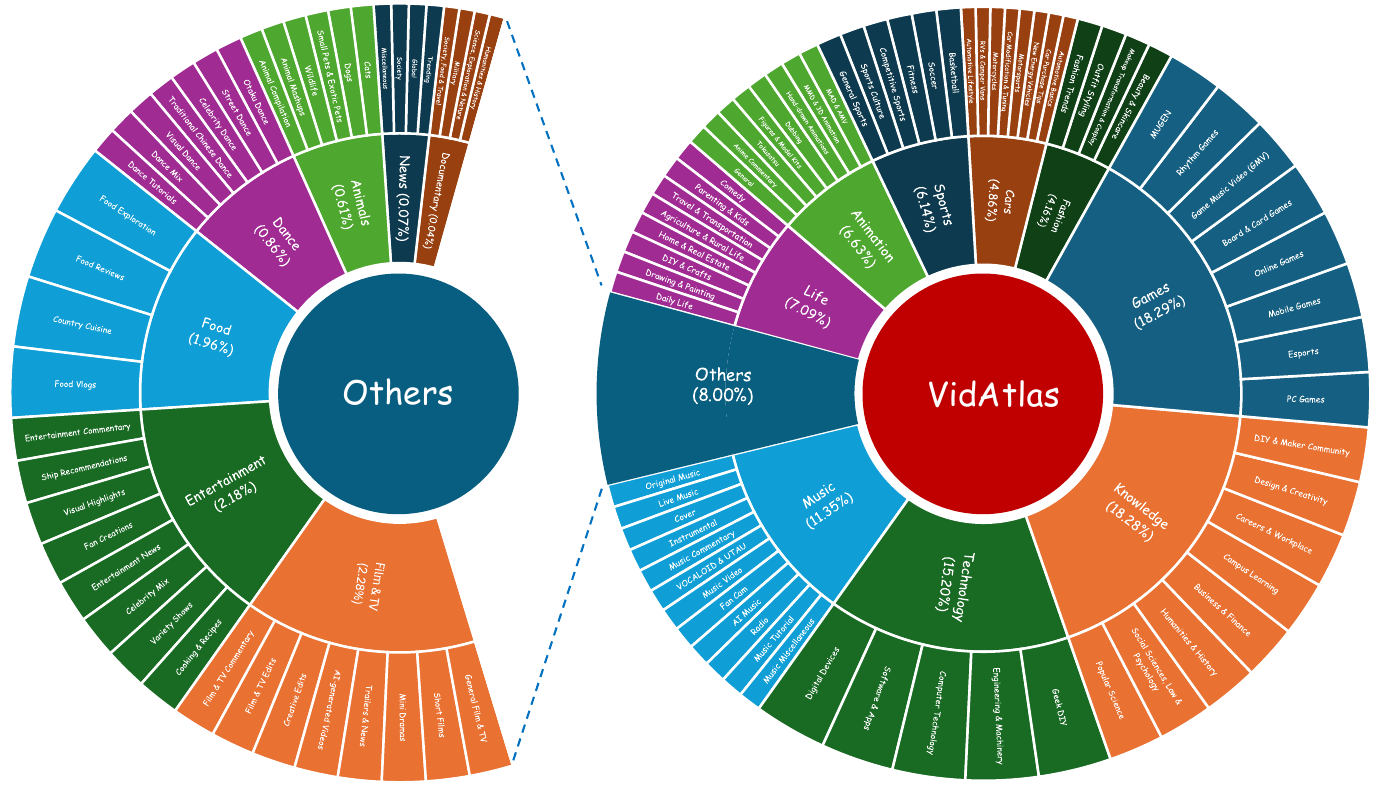}
        \caption{Categories in dataset}
        \label{fig:dataset_cat}
    \end{subfigure}

    \vspace{-2mm} % 可选：调整子图与主标题之间的垂直间距

    % --- 整个图例的总标题 ---
    \caption{\textbf{Dataset statistics:} (a) Distribution of video durations (top) and chapter durations (bottom) in the VidAtlas dataset. (b) Distribution of video topics in VidAtlas.}
    \label{fig:dataset_overview} % 为整个图例设置一个统一的标签
\end{figure}
\subsection{Hierarchical Annotation}

To generate high-quality video chaptering annotations, we design an automated annotation pipeline that leverages both multimodal content extraction and large language model (LLM)-based reasoning based on the videos with user-provided chapter makers, \ie~timestamps and brief title of each chapter. The illustration of our annotation pipeline is shown in Fig.~\ref{fig:datapipeline}.

\paragraph{\textbf{Multimodal Information Extraction.}} Considering efficiency and cost, we avoid directly using multimodal large language models (MLLMs) for video annotation. Instead, we first extract multimodal information from video frames and audio, integrate this content, and then feed the result into text-only LLM for reasoning and annotation.
Specifically, we use Whisper-v3~\cite{radford2023robust} to transcribe speech into text, segmented into sentences with the corresponding timestamps.
In parallel, we uniformly sample video frames with a fixed sampling frame rate and employ Qwen2.5-VL-7B~\cite{bai2023qwenvl} to extract visual captions and on-screen text (OCR) for better understanding of the video content.
Subsequently, the visual captions and ASR transcripts are temporally aligned based on their respective timestamps.
This process allows us to interleave the textual content from both modalities into a unified chronologically ordered sequence.
This multimodal transcript, together with the original user-provided chapter timestamps and short titles, is fed into LLM for reasoning and structural segmentation. 

\paragraph{\textbf{LLM Reasoning and Chaptering.}} The LLM is prompted to analyze the transcript and reorganize the content into a structured set of chapters, each containing a comprehensive title, an abstract, an introduction, and precise temporal boundaries.
Following this, we perform a verification step on the LLM's output to ensure that the generated chapter boundaries strictly adhere to the original timestamps.
Building upon the verified structured chapter information, we further prompt the LLM to produce a comprehensive, timestamped narrative description for the entire video.
Through this annotation pipeline, we can efficiently obtain accurate, multi-level video chapter segmentation and descriptive annotations. The resulting annotations form a dense, hierarchically organized representation of long-form videos, supporting a wide range of research tasks in video understanding, temporal reasoning, chaptering, and summarization.

\subsection{Dataset Statistics}
We summarize the key statistics of our VidAtlas dataset and highlight the properties that make it suited for research on video chaptering and summarization.
The dataset comprises 410k+ videos with an average duration of 16.8 minutes, amounting to more than 115k hours of diverse content. On average, each video is segmented into 5.5 chapters, with an average chapter duration of 182 seconds (approximately 3 minutes).
Fig.~\ref{fig:dataset_dis} provides a detailed statistic of the duration distributions for both videos and chapters. Our dataset contains a wide spectrum of video and chapter lengths to ensure models are trained on a diverse temporal structures. This comprehensive video/chapter length distribution makes the models exposed to a variety of content length, from concise segments to hour-long narratives, forcing models to resolve both rapid topic shifts and sustained thematic segments.
To mitigate genre bias, VidAtlas covers a wide array of topics, including 16 primary categories with over 100 subcategories, as shown in Fig.~\ref{fig:dataset_cat}.
The categories of VidAtlas include Games, Knowledge, Technology, Music, Life, Animation, and Sports, together with other variety that captures long-tail topics.
Videos in these categories are typically well-structured and information-dense, making them ideal for chaptering.

\section{ARC-Chapter}
\label{sec:method}

\subsection{Overall Framework}
\begin{figure}[h!]
    \centering
    \vspace{-2mm}  
    \includegraphics[width=0.99\textwidth]{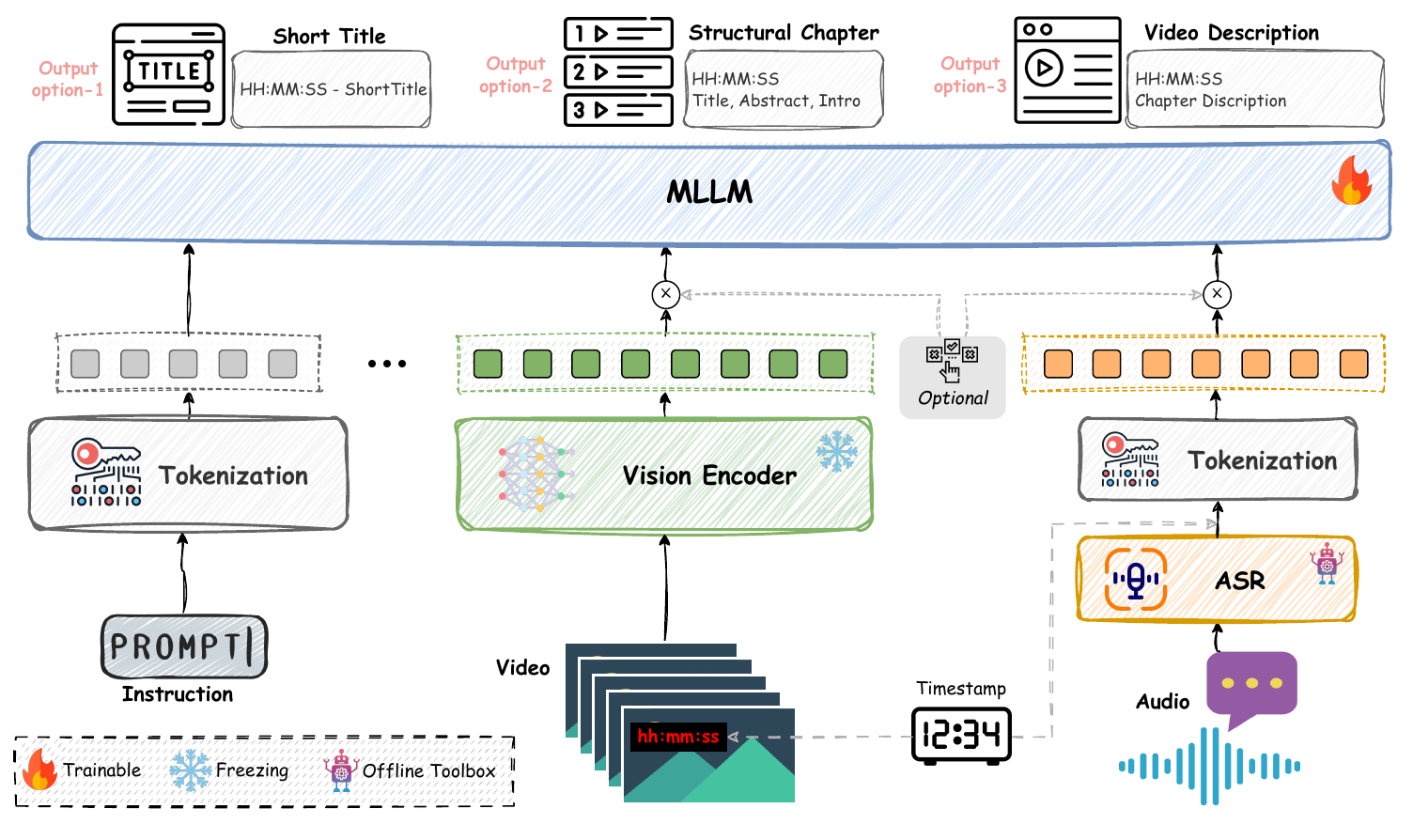}
    \caption{\textbf{Overview of the model architecture for video chaptering.} The model inputs include a task-specific prompt, sampled video frames, and timestamped ASR transcripts. Video frames are processed with a frozen vision encoder. The resulting visual features, along with the tokenized prompt and ASR text, are fed into a trainable multimodal large language model (MLLM). Based on the inputs, the model is able to generate chapters in various formats, including timestamped concise title, detailed structural chapters, or comprehensive video description with timestamps.}
  \label{fig:model} 
\end{figure}

We leverage Qwen2.5-VL-7B~\cite{bai2025qwen2} as our base model, enhancing its capabilities to process and structure video content into chapters. The architecture of our model is illustrated in Fig.~\ref{fig:model}. 
The model unifies three inputs: 1) an instruction prompt that specifies the task of input modalities and output schema. 2) a sequence of sampled video frames that provide appearance, layout and on-screen text (including subtitles which often align with the ASR transcript), and 3) a timestamp-aligned ASR transcript from audio.
While both the video and ASR transcript inputs are optional, the model requires at least one modality to be provided.
Frames are embedded with Qwen2.5-VL vision encoder and translated into visual tokens, while ASR transcript is tokenized as plain text with explicit timestamps.
The vision encoder is kept frozen and the language model is instruction tuned on VidAtlas to specialize in video chaptering.

\paragraph{\textbf{Prompt Design.}} The model's behavior is guided by carefully designed prompts that specify the desired task and output format. 
To handle the diverse requirements of different inputs and outputs of the model, we design a set of 18 distinct prompt templates.
These prompts are constructed based on three axes: language in source video, input modality, and desired output format.

\begin{itemize} 
\item \textbf{Language:} We support English and Chinese to match the language of the source video. 
\item \textbf{Input Modality:} The prompt specifies whether the model should rely on ASR-only, video-only, or both video and ASR inputs. This allows for ablation studies and adaptation to scenarios where one modality may be absent or noisy. 
\item \textbf{Output Format:} We define three distinct output structures: (a) \textit{Short Titles} for concise chapter markers, (b) \textit{Structured Chapters} that include a title, abstract, and introduction for each chapter, and (c) \textit{Video Descriptions} that provide a dense, timestamp-aligned summary of the entire video. \end{itemize}

\paragraph{\textbf{Video Input.}} To balance temporal coverage and context budget, we follow the setup of Qwen2.5‑VL and cap the visual stream at 768 frames sampled at up to 1 fps.
That is to say, videos shorter than 12.8 minutes are sampled with 1 fps, while longer videos are uniformly down-sampled to 768 frames with a lower fps.
The sampling strategy retains coarse global coverage for hour‑long content, ensuring sufficient representation to capture the high-level semantic shifts necessary for the chaptering task.
Since the model context length is shared across modalities, we dynamically adjust the per‑frame token allowance according to the input of ASR transcript.
For video-only inputs we use a higher frame resolution (higher token budget per frame) so that small text (OCR and subtitles) and fine-grained visual cues are preserved. When ASR is provided alongside video, we reduce frame resolution (thus reducing the number of visual tokens) so that the combined input of visual tokens and ASR text fits the maximum context length of MLLM. 
This dynamic allocation is implemented by adjusting image scaling and patch-tokenization parameters at preprocessing time.
Moreover, to enhance temporal awareness, we randomly overlay timestamps onto the video
frames, making the model more sensitive to the video timeline.

\paragraph{\textbf{ASR Input.}}
Although integrating raw audio features or learned audio embeddings from pretrained ASR models (\eg Whisper~\cite{radford2023robust}) is attractive, it presents severe scalability challenges for long-form video.
For example, while Whisper-style audio encoder produces 50 audio tokens per second, 
a 60-minute audio therefore produces 180k tokens, far exceeding feasible LLM context budgets without aggressive compression or specialized audio-to-token aggregation.
Furthermore, synchronizing fixed-rate audio features with dynamically sampled video frames poses an additional alignment problem.
To address these practical constraints, we opt to use ASR transcripts as a highly effective proxy for the audio modality.
Text is significantly more information-dense.
Therefore, the ASR transcript of a long audio segment occupies far fewer tokens than its raw feature representation. 
This makes processing hour-long videos computationally feasible for both training and inference.
Although such a paradigm introduces an extra step for offline ASR transcription, 
we believe that trading a modest amount of offline processing time for the ability to handle long‑form audio under strict context‑length budgets is worthwhile. 
In our implementation, we use Whisper-large-v3~\cite{radford2023robust} to generate timestamped ASR transcripts.
The model provides sentence-level segments with corresponding start timestamps.
We formulate the ASR text and timestamp of each segment as \textit{start time (hh:mm:ss): <ASR text>}.
The normalized ASR transcript is then passed to the model either alone (ASR‑only) or together with visual tokens (ASR+Video), providing dense semantic information that is particularly useful for temporal boundary detection and chaptering.

\subsection{Training Strategy}

\paragraph{\textbf{Training Objective.}}
We perform supervised instruction tuning on VidAtlas and VidChapter-7M using all prompt templates.
The training objective is the standard autoregressive next-token prediction loss over the target sequence.
Given a multimodal input sequence consisting of a prompt $X_{\rm prompt}$, video frames $X_{\rm video}$, and an ASR transcript $X_{\rm asr}$ (video stream $X_{\rm video}$ and ASR streams $X_{\rm asr}$ are optional), the model is trained to maximize the log-likelihood of the target output sequence $Y = (y_1, y_2, ..., y_n)$ (\eg, a list of chapter titles, a structured chapter object, or a timestamped description):
\vspace{-4pt}
$$ \mathcal{L} = - \sum_{i=1}^{n} \log P\left(y_i \mid y_{<n}, X_{\rm prompt}, X_{\rm video},X_{\rm asr}\right), 
\vspace{-4pt}
$$
where $y_{<i}$ represents the preceding ground-truth tokens.
During training, the vision encoder is frozen to enable a larger context length, while all parameters of the large language model are optimized with the training objective.

\paragraph{\textbf{Adaptive Modality Dropping.}}
To enable a single model to perform well under various deployment conditions, we adopt an {adaptive modality dropping} strategy during training.
For each training sample, we randomly configure the input with a certain probability to be one of three types: 1) \textbf{Video + ASR}: Both modalities are provided to the model. 2) \textbf{Video-only}: The ASR transcript is omitted, forcing the model to rely solely on visual information. and 3) \textbf{ASR-only}: The video frames are omitted, requiring the model to understand the content based on the transcript alone. 
This strategy prevents the model from becoming overly reliant on a single modality and ensures it develops a comprehensive understanding from all available input modalities.
Consequently, a single trained model can be deployed to handle videos under various conditions during inference (whether only a video is available, only transcript is provided, or both are present), without requiring specialized models for each scenario.

\subsection{Evaluation Metrics}

Evaluation metrics can be divided into two aspects: (1) the accuracy of segmentation (\eg, Precision, Recall, and tIOU~\cite{krishna2017dense}), and (2) joint metrics that assess both segmentation and chapter captioning (\eg, CIDEr~\cite{krishna2017dense}, SODA~\cite{fujita2020soda}). However, we observe that the primary metrics such as SODA, originally developed for dense video captioning, are not well-suited for the video chaptering task.
While SODA enforces a one-to-one matching between predicted and ground-truth events to suppress redundancy in overlapping event detection, video chaptering requires segmenting videos into sequential, non-overlapping chapters. Furthermore, chaptering annotations often exhibit granularity ambiguity: different annotators may segment the same video at varying levels of detail—some may annotate coarse-grained chapters (e.g., by day in a travel vlog), while others may provide fine-grained chapters (e.g., by each visited site within a day). This results in multiple valid annotation granularities for the same content.

\begin{figure}[h]
    \centering
    \begin{subfigure}[b]{0.45\textwidth}
        \centering
        \caption{One-to-One Matching: SODA}
        \begin{tikzpicture}[scale=0.8]
            % Panel (a): One-to-One Matching
            
            % Prediction line and markers
            \draw[predblue, very thick] (0,3) -- (8,3);
            \node[mydiamond=predblue] at (0,3) {};
            \node[mydiamond=predblue] at (5,3) {};
            \node[mydiamond=predblue] at (6,3) {};
            \node[mydiamond=predblue] at (8,3) {};
            
            % Prediction labels
            \node[textgray, above] at (0,3.3) {\textbf{Pred}};
            \node[textgray, above] at (2.5,3.3) {$p_1$};
            \node[textgray, above] at (5.5,3.3) {$p_2$};
            \node[textgray, above] at (7,3.3) {$p_3$};
            
            % Ground truth line and markers
            \draw[gtgreen, very thick] (0,1) -- (8,1);
            \node[mydiamond=gtgreen] at (0,1) {};
            \node[mydiamond=gtgreen] at (2,1) {};
            \node[mydiamond=gtgreen] at (4,1) {};
            \node[mydiamond=gtgreen] at (8,1) {};
            
            % Ground truth labels
            \node[textgray, below] at (0,0.7) {\textbf{GT}};
            \node[textgray, below] at (1,0.7) {$g_1$};
            \node[textgray, below] at (3,0.7) {$g_2$};
            \node[textgray, below] at (6,0.7) {$g_3$};
            
            % Matching lines (one-to-one)
            \draw[matchgray, dashed, thick] (2.5,2.8) -- (1,1.2);
            \draw[matchgray, dashed, thick] (7,2.8) -- (6,1.2);

            % Formula box
            \node[textgray, fill=gray!10, rounded corners, minimum height=0.8cm, minimum width=6cm] 
                at (4,-5mm) {$\mathrm{SODA} = \lambda_{1} \mathrm{Sim}(p_1, g_1) + \lambda_{2} \mathrm{Sim}(p_3, g_3)$};
            
        \end{tikzpicture}
        \label{fig:soda}
    \end{subfigure}
    % \hfill
    \ 
    \begin{subfigure}[b]{0.5\textwidth}
        \centering
        \caption{Many-to-One Matching: GRACE}
        
        \begin{tikzpicture}[scale=0.8]
            % Panel (b): One-to-Many Matching
            
            % Prediction line and markers
            \draw[predblue, very thick] (0,3) -- (8,3);
            \node[mydiamond=predblue] at (0,3) {};
            \node[mydiamond=predblue] at (5,3) {};
            \node[mydiamond=predblue] at (6,3) {};
            \node[mydiamond=predblue] at (8,3) {};
            
            % Prediction labels
            \node[textgray, above] at (0,3.3) {\textbf{Pred}};
            \node[textgray, above] at (2.5,3.3) {$p_1$};
            \node[textgray, above] at (5.5,3.3) {$p_2$};
            \node[textgray, above] at (7,3.3) {$p_3$};
            
            % Ground truth line and markers
            \draw[gtgreen, very thick] (0,1) -- (8,1);
            \node[mydiamond=gtgreen] at (0,1) {};
            \node[mydiamond=gtgreen] at (2,1) {};
            \node[mydiamond=gtgreen] at (4,1) {};
            \node[mydiamond=gtgreen] at (8,1) {};
            
            % Ground truth labels
            \node[textgray, below] at (0,0.7) {\textbf{GT}};
            \node[textgray, below] at (1,0.7) {$g_1$};
            \node[textgray, below] at (3,0.7) {$g_2$};
            \node[textgray, below] at (6,0.7) {$g_3$};
            
            % Matching lines (one-to-many, curved)
            \draw[matchgray, dashed, thick] (2.5,2.8) to[out=270, in=90] (1,1.2);
            \draw[matchgray, dashed, thick] (2.5,2.8) to[out=270, in=90] (3,1.2);
            \draw[matchgray, dashed, thick] (5.5,2.8) to[out=270, in=90] (6,1.2);
            \draw[matchgray, dashed, thick] (7,2.8) to[out=270, in=90] (6,1.2);
            
            % Formula box
            \node[textgray, fill=gray!10, rounded corners, minimum height=0.8cm, minimum width=6cm] 
                at (4,-5mm) {$\mathrm{GRACE} = \varphi_{1} \mathrm{Sim}(p_1, g_1 \cup g_2) + \varphi_{2} \mathrm{Sim}(p_2 \cup p_3, g_3)$};
            
        \end{tikzpicture}
        \label{fig:grace}
    \end{subfigure}
     \vspace{-6pt}
     \caption{
     \textbf{Comparison of one-to-one (SODA) and many-to-one (GRACE) matching strategies.} The one-to-one matching can fail to account for important events like $p_2$ and $g_2$, whereas the many-to-one strategy considers all predicted and ground-truth events for a more robust, overall assessment.
    }
    \label{fig:metric}
    \vspace{-8pt}
\end{figure}
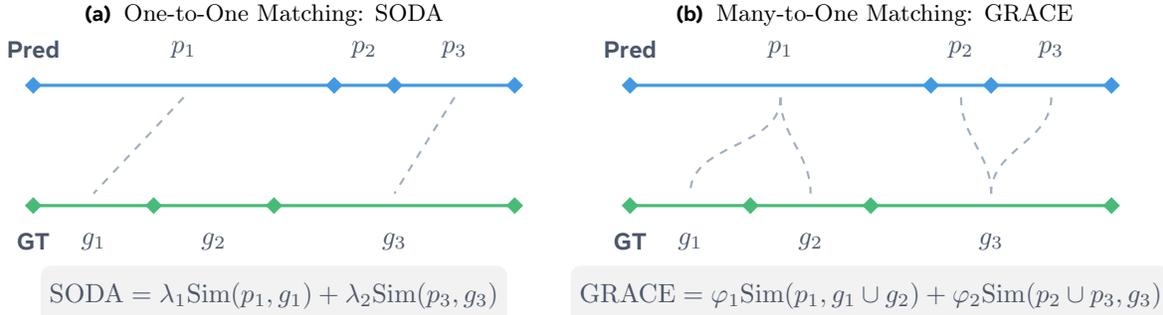

To address these challenges, we propose GRACE, a metric tailored for video chaptering. It introduces a many-to-one (set-to-one) matching paradigm, allowing each ground-truth (predicted) chapter to be matched with a set of predicted (ground-truth) chapters. As illustrated in Fig.~\ref{fig:metric}, for each ground-truth chapter, GRACE evaluates the temporal overlap and semantic similarity between the chapter and its matched prediction set, using established language similarity metrics (e.g., BERTscore~\cite{zhang2019bertscore}) for textual comparison. Specifically, we aim to find a best many-to-one mapping $M$ which splits both ground-truth set $G$ and prediction set $P$ into several pairs of groups $\{(P_i, G_i)\}^{K}_{i=1}$, followed by group-based similarity calculation:
\vspace{0pt}
\begin{align}
    &{\rm GRACE}=\sum_{(P_{i}, G_{i})\in M(P, G)} \ {\rm \varphi}(P_i, G_i) \cdot {\rm BERTscore}(P_i, G_i) \label{eq:grace} \\
    &{\rm \varphi}(P_i, G_i) = \frac{1}{|P_i||G_i|}\sum_{p\in P_i, g\in G_i} {\rm IOU}(p, g) \\
    &\textbf{s.t.} \ P_i \cap P_j = \emptyset, \ \cup ({P_i}) = P, \ G_i \cap G_j = \emptyset, \ \cup ({G_i}) = G , \  \min(|P_i|, |G_i|)=1
\end{align}
where $P_i$ and $G_i$ epresent groups of chapters. When calculating the BERTScore between two groups, we first concatenate all captions within each group into a single sentence, then compute the BERTScore between the two merged sentences. We adopt the dynamic time warping algorithm (DTW)~\cite{bellman2003adaptive, sakoe2003dynamic} to achieve the optimal matching $M(P, G)$, with IOU between two chapters being used as the matching criteria. 

GRACE provides a more accurate and human-aligned assessment of chaptering models. This design confers several advantages: (1) robustness to annotation granularity, enabling fair evaluation across diverse annotation styles; (2) improved semantic fidelity, rewarding models that capture the full scope of ground-truth chapters; and (3) closer alignment with human judgment of chapter boundaries and content. 

\subsection{Reinforcement Learning with GRPO}
While supervised fine-tuning (SFT) achieves strong performance, the standard cross-entropy loss does not directly optimize for the primary objective of video chaptering: temporal accuracy.
To further enhance the model's temporal localization capabilities, we introduce a subsequent reinforcement learning phase using the GRPO algorithm~\cite{guo2025deepseek}.

The core of this phase is a reward function designed to directly incentivize precise chapter boundary prediction. We leverage our proposed GRACE metric, which holistically evaluates both temporal alignment and semantic content. However, to specifically sharpen the model's ability to predict accurate timestamps of segmented chapters, we formulate a simplified, temporal-only reward by omitting the semantic BERTscore component from Equation~\eqref{eq:grace}.
For a given ground-truth chapter set $G$ and a model-generated set $P$, the reward $R$ is calculated by summing the temporal alignment scores $\varphi$ over the optimal matching $M(P,G)$ found via DTW:
\vspace{-4pt}
\begin{equation}
    R =\sum_{(P_{i}, G_{i})\in M(P, G)} \ {\rm \varphi}(P_i, G_i). \label{eq:reward}
    \vspace{-4pt}
\end{equation}
This reward directly reflects the quality of the temporal segmentation, providing a clear and targeted optimization objective.

Due to the significant context length required for multimodal inputs, and to specifically bolster the model's ability to reason from visual cues, we conduct this RL training phase using only the video modality.
We select a diverse subset of 90k videos from both Chinese and English SFT data, ensuring that training samples cover all three output formats: short titles, structural chapters, and timestamped video description.
We initialize the model with the weights from our best-performing SFT model and further optimize it using GRPO.
The KL divergence coefficient is set to $0.01$ to ensure that the policy does not stray far from the robust language generation capabilities  learned during SFT, thereby balancing temporal refinement with descriptive quality.

\begin{table*}[t]
\centering
\caption{\textbf{Comparison to the state of the art on VidChapters7M-test set:} The results of compared methods are evaluated in the ASR‑only setting from Chapter-Llama~\cite{ventura2025chapter}. We evaluate ARC-Chapter with different input modalities: \textbf{-vid} for video, \textbf{-asr} for ASR, and \textbf{-vidasr} for both. ``\textbf{\textit{Ft.}}'' indicates whether the model is finetuned for chaptering task. \dag denotes LLM-API results reported from Chapter-Llama. Our model, ARC-Cchapter, achieves the best performance across all metrics and video durations.}
\label{tab:comparison_vidchaptext}
% Use resizebox to fit the table to the text width, useful for wide tables
\resizebox{\textwidth}{!}{%
\begin{tabular}{lc|cccc|cccc|cccc|cccc}
\toprule
\multirow{2}{*}{\textbf{Backbone}} & \multirow{2}{*}{\textbf{Ft.}} & \multicolumn{4}{c|}{\textbf{Short}} & \multicolumn{4}{c|}{\textbf{Medium}} & \multicolumn{4}{c|}{\textbf{Long}} & \multicolumn{4}{c}{\textbf{All}} \\
\cmidrule(lr){3-6} \cmidrule(lr){7-10} \cmidrule(lr){11-14} \cmidrule(lr){15-18}
& & F1 & tIoU & S & C & F1 & tIoU & S & C & F1 & tIoU & S & C & F1 & tIoU & S & C \\
\midrule
\midrule
GPT-4o-mini~\cite{hurst2024gpt}\dag & \ding{55} & 32.1 & 64.5 & 7.2 & 42.4 & 30.5 & 62.3 & 6.1 & 30.6 & 28.0 & 61.0 & 6.0 & 27.3 & 31.2 & 63.6 & 6.8 & 37.8 \\
GPT-4o~\cite{hurst2024gpt}\dag & \ding{55} & 37.7 & 68.0 & 8.4 & 53.8 & 38.1 & 68.8 & 8.1 & 51.4 & 36.5 & 66.2 & 6.6 & 34.8 & 37.6 & 68.0 & 8.1 & 51.0 \\
Gemini-2.0-Flash~\cite{team2023gemini}\dag & \ding{55} & 39.9 & 69.2 & 12.0 & 72.8 & 43.8 & 71.4 & 11.2 & 70.3 & 34.9 & 66.2 & 9.0 & 51.6 & 40.2 & 69.3 & 11.4 & 69.7 \\
Gemini-1.5-Pro~\cite{team2023gemini}\dag & \ding{55} & 41.7 & 70.6 & 11.7 & 65.3 & 43.8 & 71.8 & 11.2 & 61.4 & 41.3 & 70.6 & 10.1 & 55.3 & 42.2 & 70.9 & 11.4 & 63.2 \\
Vid2Seq~\cite{yang2023vidchapters,yang2023vid2seq} & \ding{55} & 2.5 & 28.6 & 0.3 & 0.3 & 3.2 & 29.7 & 0.3 & 0.4 & 4.6 & 32.0 & 0.3 & 0.5 & 3.0 & 29.3 & 0.3 & 0.4 \\
Llama 3.1-8B~\cite{dubey2024llama3} & \ding{55} & 29.9 & 63.4 & 7.1 & 34.5 & 30.6 & 62.7 & 5.4 & 28.1 & 26.6 & 59.3 & 3.6 & 18.9 & 29.5 & 62.5 & 6.2 & 30.7 \\
\midrule
Vid2Seq~\cite{yang2023vidchapters,yang2023vid2seq} & \ding{51} & 33.4 & 63.7 & 15.2 & 74.9 & 19.0 & 53.3 & 7.5 & 31.9 & 16.7 & 50.8 & 5.9 & 28.4 & 26.7 & 58.6 & 11.6 & 55.8 \\
Chapter-Llama~\cite{ventura2025chapter} & \ding{51} & 45.5 & 72.2 & 20.2 & 103.5 & 46.7 & 72.3 & 18.8 & 98.7 & 41.3 & 69.2 & 15.8 & 91.2 & 45.3 & 71.8 & 19.3 & 100.9 \\
\midrule
\textbf{ARCChapter-asr}\footnotemark & \ding{51} & {54.5} & {76.7} & {26.3} & {144.1} & {55.9} & {77.5} & {25.1} & {143.0} & {55.1} & {77.0} & {24.8} & {158.0} & {54.5} & {76.7} & {25.3} & {144.0} \\
\textbf{ARCChapter-vid} & \ding{51} &  52.6 & 75.8 & 26.0 & 156.8  &  51.4 & 75.3 & 20.6 & 124.0  &  47.3 & 72.3 & 19.2 & 119.8  &  50.2 & 74.3 & 22.9 & 138.3  \\
\textbf{ARCChapter-vidasr} & \ding{51} & \textbf{60.0} & \textbf{80.1} & \textbf{32.5} & \textbf{195.7} & \textbf{59.2} & \textbf{79.4} & \textbf{29.6} & \textbf{177.3} & \textbf{60.2} & \textbf{79.9} & \textbf{29.2} & \textbf{190.3} & \textbf{59.3} & \textbf{79.6} & \textbf{30.6} & \textbf{186.6} \\ 
\bottomrule
\end{tabular}
}
\end{table*}
\footnotetext{For convenience, "ARC-Chapter" in the main text is abbreviated as "ARCChapter" in all experimental result tables.}

\begin{table}[h!]
\centering
% \small 
\footnotesize
\setlength{\tabcolsep}{1.5pt} 
\renewcommand{\arraystretch}{1.0} % 增加行高以获得更好的视觉效果
% \begin{tabular}{lc |ccc|cc|cc}
\caption{\textbf{Comparison to the state of the art on VidChapter7M-sml300 with different input modalities.} Our method, ARC-Chapter, demonstrates superior performance on VidChapter-sml300 by effectively integrating both speech and video information. The modalities of ``Embed'' and ``Caption'' in LLaMA and Chapter-LLaMA models play the same role as ``Video'' in ARC-Chapter model.}

\resizebox{0.75\textwidth}{!}{%
\begin{tabular}{lc |ccc|>{\centering\arraybackslash}p{1.2cm}>{\centering\arraybackslash}p{1.2cm}|>{\centering\arraybackslash}p{1.2cm}>{\centering\arraybackslash}p{1.2cm}}
\toprule
\multirow{2}{*}{\textbf{Method}} & \multirow{2}{*}{\textbf{Ft.}} & \multicolumn{3}{c|}{\textbf{Modalities}} & \multicolumn{2}{c|}{\textbf{Segmentation}} & \multicolumn{2}{c}{\textbf{Titles}} \\ 
% \cline{2-7}
\cmidrule(lr){3-5} \cmidrule(lr){6-7} \cmidrule(lr){8-9}
& & Speech  & Embed.      & Caption     & F1          & tIoU         & S           & C          \\ 
% \hline\hline
\midrule
\multirow{3}{*}{\rotatebox{0}{LLaMA 3.1-8B}}
& \xmark & \ding{55}     &   \xmark    & \ding{51}          & 12.6                 & 48.6                  & 1.9                  & 6.4                 \\
& \xmark & \ding{51}    &  \xmark   & \ding{55}             & 22.7                 & 57.3                  & 4.4                  & 19.7                \\
& \xmark & \ding{51}     &  \xmark  & \ding{51}          & 29.9                 & 63.0                  & 6.9                  & 33.7                \\ 
\midrule
\multirow{6}{*}{\rotatebox{0}{Chapter-LLaMA}} 
& \cmark & \cmark    &  \xmark   & \xmark              & 38.5                 & 68.1                  & 13.9                 & 67.3                \\
& \cmark & \xmark    &  \cmark   & \xmark              & 38.4                 & 66.5                  & 3.4                 & 7.3                \\
& \cmark & \xmark   &  \xmark  & \cmark       & 39.1            & 67.7                  & 5.9                  & 20.2         \\
% \cmidrule(lr){3-5} \cmidrule(lr){6-7} \cmidrule(lr){8-9}
& \cmark & \cmark    &  \cmark   & \xmark          & 40.4        & 68.2         & 15.3        & 74.9      \\ 
& \cmark & \cmark    &  \xmark   & \cmark          & 42.6        & 70.6         & 16.4        & 82.4      \\ 
& \cmark & \cmark    &  \cmark   & \cmark          & 44.4        & 71.5         & 16.3        & 84.2      \\ 
\midrule
\multirow{4}{*}{\rotatebox{0}{\textbf{ARCChapter}}} &  & Speech  & \multicolumn{2}{c|}{Video}     & F1          & tIoU         & S           & C          \\ 
% \cmidrule(lr){3-9}
\cmidrule(lr){3-5} \cmidrule(lr){6-7} \cmidrule(lr){8-9}
& \cmark & \cmark    &     \multicolumn{2}{c|}{\xmark}         & 56.5        & 78.1         & 25.9        & 148.5       \\ 
& \cmark & \xmark    &     \multicolumn{2}{c|}{\cmark}         & 50.0        & 74.3         & 21.6        & 130.8       \\ 
& \cmark & \cmark    &     \multicolumn{2}{c|}{\cmark}         & \textbf{62.4}        & \textbf{81.6}         & \textbf{30.1}        & \textbf{190.7}       \\ 
\bottomrule

\end{tabular}
}
\label{tab:sml300}
\end{table}

\section{Experiments}
\label{sec:experiment}
In this section, we conduct a series of experiments to thoroughly evaluate our video chaptering model. We first introduce the evaluation benchmarks, then present the main results and detailed ablation studies.

\subsection{Evaluation Benchmark}
To comprehensively assess our model's capabilities in video chaptering, we evaluate it on three distinct benchmarks covering different languages, scales, and data modalities. The evaluation targets two key criteria: the precision of temporal boundary localization and semantic relevance of the generated chapter titles/descriptions. VidChapters7M is a large-scale English chaptering dataset. We use two of its standard splits for evaluation, \ie, VidChapters7M-test and VidChapters7M-sml300val.
VidChapters7M-test is a large-scale test set comprising 8.2k samples. 
For this split, the compared methods are only based on ASR transcripts, while ARC-Chapter is evaluated with different input modalities.
VidChapters7M-sml300val is a smaller validation set of 300 samples, which includes both the original videos and their corresponding ASR transcripts. This subset is ideal for fast evaluation and conducting modality ablation studies.
To assess generalization beyond English, we additionally report experimental results on VidAtlas‑test, a Chinese test set with more than 1.5k videos together with ASR transcripts and original videos.

\subsection{Comparison with the State of the Art}
\paragraph{\textbf{Performance on VidChapters7M.}}
As shown in Tab.~\ref{tab:comparison_vidchaptext}, our ARC-Chapter significantly outperforms all existing methods on VidChapters7M-test benchmark.
Our model achieves a new state-of-the-art result in the ASR‑only regime, with an overall F1 score of 54.5, tIoU of 76.7, SODA of 23.5, and a CIDEr of 144.0. 
This represents a substantial improvement over the previous SOTA model, Chapter-Llama, with absolute gains of +9.2 in F1, +4.9 in tIoU, and +6.0 in the SODA score. 
Notably, the performance gain enlarges as video duration increases. 
For long videos (30-60 min), the evaluation metrics of SODA and CIDEr for ARC-Chapter are remarkably higher than which in Chapter-LLama, demonstrating the superior capability of our model in processing long videos.
Even when compared against powerful general models like GPT-4o and Gemini-1.5-Pro, which are not finetuned on this task, ARC-Chapter perform much better.
The experiments conducted on VidChapter7M-sml300 show more comparisons for different input modalities, shown in Tab.~\ref{tab:sml300}.

\paragraph{\textbf{Performance on VidAtlas.}} 
As detailed in Tab.~\ref{tab:compare_cnbench_test}, we evaluate our model on the VidAtlas benchmark under three settings: ASR-only, video-only, and ASR$+$video. 
ARC-Chapter consistently establish a new state-of-the-art across all settings.
Our full multimodal model, ARCChapter-vidasr, which leverages both ASR and video inputs, achieves an overall F1 score of 66.2, tIoU of 84.0, SODA of 30.2, CIDEr of 141.5, and GRACE of 34.1. This marks a significant leap over the strongest LLM, Gemini-2.5-Pro, with an absolute improvement of +17.5 in F1 score and more than doubling the SODA score (+16.7). Furthermore, our single-modality versions also demonstrate superior performance. The ASR-only model, ARCChapter-asr, achieves an F1 of 58.8, and the video-only model, ARCChapter-vid, scores an F1 of 57.6. From shot-to-long videos, our model consistently outperforms other models, demonstrating its robustness in handling extended content.

\begin{table*}[t]
\centering
\caption{\textbf{Comparison to the state of the art on VidAtlas-test set:} ``\textbf{\textit{Ft.}}'' indicates whether the model is finetuned for chaptering task. Modality\ddag~specifies which inputs are provided: A for ASR and V for video. \dag~denotes LLM-API results. For API-base models, the video is converted into a textual description, which is then provided as input for LLM.}
\vspace{-3pt}
\label{tab:compare_cnbench_test}
% Use resizebox to fit the table to the text width, useful for wide tables
\resizebox{\textwidth}{!}{%
\begin{tabular}{lc|cc|cccc|cccc|cccc|ccccc}
\toprule
\multirow{2}{*}{\textbf{Backbone}} & \multirow{2}{*}{\textbf{Ft.}} & \multicolumn{2}{c|}{\textbf{Modality}\ddag} & \multicolumn{4}{c|}{\textbf{Short}} & \multicolumn{4}{c|}{\textbf{Medium}} & \multicolumn{4}{c|}{\textbf{Long}} & \multicolumn{5}{c}{\textbf{All}} \\
\cmidrule(lr){3-4} \cmidrule(lr){5-8} \cmidrule(lr){9-12} \cmidrule(lr){13-16} \cmidrule(lr){17-21}
& & \textbf{A} & \textbf{V} & F1 & tIoU & S & C & F1 & tIoU & S & C & F1 & tIoU & S & C & F1 & tIoU & S & C & G \\
\midrule
\midrule

Claude-Sonnet~\cite{anthropic2024claude}\dag & \xmark & \cmark & \xmark& 39.2 & 69.8 & 7.6 & 38.8 &     34.7 & 66.3 & 6.5 & 33.8 &     36.6 & 66.9 & 5.8 & 33.5 &     37.8 & 68.6 & 7.1 & 36.9 & 11.1 \\
Doubao-1.5-Pro~\cite{guo2025seed1}\dag & \xmark & \cmark & \xmark& 38.8 & 70.4 & 7.4 & 40.6 &     35.8 & 68.4 & 6.9 & 38.3 &     36.1 & 67.1 & 3.2 & 17.4 &     37.7 & 69.5 & 6.7 & 36.4 & 9.8 \\
DeepSeek-R1~\cite{guo2025deepseek}\dag & \xmark & \cmark & \xmark& 40.0 & 71.1 & 11.0 & 48.8 &     37.9 & 69.5 & 9.6 & 45.2 &     35.7 & 66.8 & 6.3 & 28.3 &     38.9 & 70.1 & 10.0 & 44.8 & 13.4 \\
Gemini-2.5-Pro~\cite{comanici2025gemini}\dag & \xmark & \cmark & \xmark& 39.6 & 68.3 & 8.1 & 44.6 &     30.6 & 60.1 & 6.3 & 37.4 &     34.0 & 60.2 & 9.9 & 54.0 &     45.2 & 73.2 & 9.7 & 53.5 & 14.9 \\
GPT-4.1~\cite{achiam2023gpt}\dag & \xmark & \cmark & \xmark& 36.5 & 68.6 & 6.6 & 34.6 &     33.0 & 66.1 & 5.8 & 32.4 &     36.0 & 66.3 & 5.9 & 33.0 &     35.7 & 67.7 & 6.3 & 33.9 & - \\
Qwen3-235B~\cite{yang2025qwen3}\dag & \xmark & \cmark & \xmark& 36.7 & 67.7 & 7.7 & 36.9 &     33.5 & 65.6 & 6.7 & 33.9 &     26.6 & 61.0 & 3.8 & 18.7 &     34.4 & 66.2 & 6.9 & 33.4 & 10.2 \\

\midrule

Claude-Sonnet~\cite{anthropic2024claude}\dag & \xmark & \cmark & \cmark & 36.8 & 68.2 & 7.9 & 42.4 &     32.0 & 65.2 & 8.0 & 45.0 &     40.8 & 68.2 & 16.8 & 110.4 &     36.4 & 67.5 & 9.3 & 53.6 & 13.2 \\
Doubao-1.5-Pro~\cite{guo2025seed1}\dag & \xmark & \cmark & \cmark& 39.5 & 70.0 & 7.7 & 43.3 &     35.5 & 67.6 & 7.6 & 45.2 &     44.4 & 69.8 & 14.9 & 109.0 &     39.5 & 69.4 & 8.8 & 54.1 & 12.6\\
DeepSeek-R1~\cite{guo2025deepseek}\dag & \xmark & \cmark & \cmark& 39.4 & 69.9 & 10.5 & 50.0 &     38.0 & 68.7 & 10.8 & 54.9 &     \underline{62.2} & \underline{80.3} & \underline{48.2} & \underline{264.4} &     41.1 & 70.5 & 13.9 & 69.7 & 17.1 \\
Gemini-2.5-Pro\cite{comanici2025gemini}\dag & \xmark & \cmark & \cmark & 48.3 & 73.1 & 9.8 & 54.9 &     45.4 & 70.1 & 11.8 & 66.1 &     54.8 & 75.3 & 30.6 & 172.5 &     48.7 & 72.8 & 13.5 & 75.8 & 19.8 \\
GPT-4.1~\cite{achiam2023gpt}\dag & \xmark & \cmark & \cmark & 35.3 & 67.2 & 6.3 & 34.2 &     30.8 & 64.2 & 6.2 & 34.8 &     43.9 & 69.2 & 19.1 & 120.2 &     35.8 & 66.9 & 8.3 & 47.9 & 11.7\\
Qwen3-235B~\cite{yang2025qwen3}\dag & \xmark & \cmark & \cmark & 24.8 & 59.2 & 6.5 & 31.9 &     19.5 & 52.9 & 5.6 & 28.2 &     27.5 & 57.8 & 16.0 & 92.9 &     24.1 & 57.7 & 7.8 & 40.7 & 9.6\\

\midrule

\textbf{ARCChapter-asr} & \cmark & \cmark & \xmark & {57.3} & {79.3} & {24.1} & {103.3} & {60.1} & {80.8} & {24.5} & {113.5} & {63.2} & {79.5} & {28.1} & {140.6} & {58.8} & {79.7} & {24.8} & {111.3} & 28.0 \\

\textbf{ARCChapter-vid} & \cmark & \xmark & \cmark & {57.1} & {79.1} & {21.2} & {91.5} & {55.9} & {78.2} & {18.4} & {88.2} & {62.0} & {79.4} & {27.9} & {137.8} & {57.6} & {78.9} & {21.6} & {98.1} & 25.0 \\

\textbf{ARCChapter-vidasr} & \cmark & \cmark & \cmark & \textbf{65.5} & \textbf{83.8} & \textbf{28.5} & \textbf{129.2} & \textbf{65.7} & \textbf{84.2} & \textbf{29.0} & \textbf{140.0} & \textbf{69.6} & \textbf{84.2} & \textbf{38.5} & \textbf{192.3} & \textbf{66.2} & \textbf{84.0} & \textbf{30.2} & \textbf{141.5} & \textbf{34.1} \\
\bottomrule
\end{tabular}
}
\end{table*}
% \vspace{-0pt}

\subsection{Transferability}

\begin{table*}[t]
    \centering
    \small
    \caption{\textbf{Transferability Performance on YouCook2 and ActivityNet Captions \cite{krishna2017dense} for Dense Video Captioning. } All methods use visual modality as inputs without ASR.  The \textbf{Rank}($\downarrow$) column represents the overall performance, calculated as the arithmetic mean of a method's rank across all reported metrics (M, S, C, and F1) for that dataset. Some results for ActivityNet Captions are sourced from \cite{guo2024trace} and \cite{yang2023vid2seq}. * indicates zero-shot evaluation. The best results on each dataset are in \textbf{bold} and the second-best are \underline{underlined}.}
    \label{tab:dvc_results_merged_highlighted}

    \setlength\tabcolsep{6pt}
    \renewcommand{\arraystretch}{1.0}
    
    % You might need to add \usepackage{multirow}, \usepackage{colortbl}, \usepackage{booktabs} to your preamble
    \begin{tabular}{lcccccccccc}
        \toprule
        \multirow{2}{*}{\textbf{Method}} & \multicolumn{5}{c}{\textbf{YouCook2}} & \multicolumn{5}{c}{\textbf{ActivityNet Captions}} \\
        \cmidrule(lr){2-6} \cmidrule(lr){7-11}
        & M & S  & C & F1  &\textbf{Rank}$\downarrow$& M & S & C & F1 & \textbf{Rank}$\downarrow$ \\
        \midrule
        % Traditional Models
        GIT \cite{wang2022git} & 3.4 & 3.1 & 12.1 & 17.7 & 7.5 
        & 7.8 & 5.7 & 29.8  & 50.6 & 4.3\\
        ECHR \cite{yang2023vid2seq} & 3.8 & - & - & - & 4.0 
        & 7.2 & 3.2 & 14.7 & - & 8.6\\
        PDVC \cite{yang2023vid2seq} & 4.7 & 4.4 & 22.7 & - & 5.0 
        & 8.0 & 5.4 & 29.0 & \textbf{56.7} & 3.8\\
        Vid2Seq \cite{yang2023vid2seq} & \underline{9.3} & \underline{7.9} & \underline{47.1} & 27.3 & {2.8}
        & \underline{8.5} & 5.8 & \underline{30.1} & 52.4 & \underline{2.6} \\
        CM$^2$ & - & 5.3 & 31.7 & 28.4 & 4.7 
        & - & - & - & - & - \\
        \midrule
        % VTG-specific Video-LLMs
        TimeChat \cite{ren2024timechat} & - & 3.4 & 11.0 & 19.5 & 8.0 
        & 5.7 & 4.7 & 19.0 & 36.9 & 8.8\\
        VTimeLLM \cite{huang2024vtimellm} & - & - & - & - & - & 6.8 & 5.8 & 27.6 & - & 5.8\\
        Momentor$^{*}$ \cite{qian2024momentor} & - & - & - & - &- &  4.7 & 2.3 & 14.9 & - & 10.7 \\
        TRACE \cite{guo2024trace} & - & 6.7 & 35.5 & 31.8 & 3.7
        & 6.4 & \underline{6.0} & 25.9 & 39.3 & 5.8 \\
        VTG-LLM \cite{guo2025vtg} & - & 3.6 & 13.4 & 20.6 & 6.7 
        & 5.9 & 5.1 & 20.7 & 34.8 & 8.3 \\
        TimeExpert \cite{yang2025timeexpert} & - & {7.2} & {39.0} & {33.5} & \underline{2.7}
        & 7.0 & \textbf{6.5} & 28.4 & 40.5 & 4.3 \\
        % \midrule
        % Main method and baselines
        \rowcolor{gray!20}
        \textbf{ARC-Chapter} & \textbf{9.6} & \textbf{12.5} & \textbf{69.4} & \textbf{37.9} & \textbf{1.0}
        & \underline{8.1} & 5.9 & \textbf{{35.4}} & \underline{55.9} & \textbf{2.0} \\
        \bottomrule
    \end{tabular}
\end{table*}

To evaluate transferability, we pre-trained ARC-Chapter on our dataset before fine-tuning and testing it on the dense video captioning benchmarks, \ie, Youcook2 and ActivityNet Captions. As shown in Table~\ref{tab:dvc_results_merged_highlighted}, our model establishes a new state-of-the-art, significantly outperforming all prior MLLM-based methods. 

Notably, for event segmentation ability, ARC-Chapter achieves an F1/SODA Score of 37.9/12.5 on YouCook2, a substantial improvement over the previous best of 33.5/7.9. This demonstrates that the knowledge acquired during pre-training effectively transfers and enhances performance on downstream tasks.

\begin{figure}
    \centering
    \includegraphics[width=1.0\textwidth]{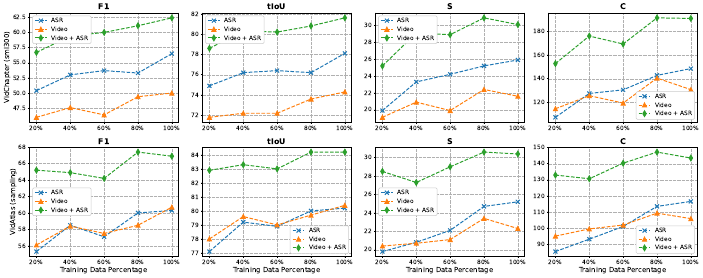}
    \vspace{-3mm}  
    \caption{\textbf{Data Scaling property of ARC-Chapter.} We report the performance on VidChapter (a sampled subset) and VidAtlas test set with respect to different percentage of training samples.}
  \label{fig:data_volume} 
\end{figure}

\subsection{Ablation Studies}

\subsubsection{Scaling Property}
We analyze how ARC-Chapter scales with the amount of training data.
Concretely, we subsample the training set at 20\%, 40\%, 60\%, 80\%, and 100\% and keep the model architecture and prompt templates fixed.
We evaluate three inference modalities, \ie ASR-only, Video-only, and ASR+Video, on two benchmarks: VidChapters-7M (sml300val) and a sampled subset of the VidAtlas-testset for efficiency.
As illustrated in Fig.~\ref{fig:data_volume}, the performance across all metrics (F1, tIOU, SODA, and CIDEr) and input modalities (ASR-only, Video-only, Video+ASR) demonstrates a clear positive correlation with the amount of training data.
Specifically, the full multimodal model (Video+ASR) consistently achieves the best performance.
ARC-Chapter is highly data-efficient, achieving strong performance with as little as 20\% of the training data.
Furthermore, it is data-scalable, continuing to benefit from larger corpora for even better results.

\subsubsection{Hierarchical Annotations}
A core contribution of our work is the VidAtlas dataset, which features rich, hierarchical annotations.
To validate the effectiveness of this data structure, we evaluate our model's capability to generate outputs of varying complexity, from simple \textit{Short Title} to detailed \textit{Structural Info} which comprising a title, abstract and introduction for each chapter.
The results are presented in Table~\ref{tab:hierarchical_output_cnbench}.
From the experimental results, our model successfully learns to generate these complex, structured outputs, achieving strong performance across all generated components (title, abstract, introduction) on both VidChapter-sml300 and VidAtlas-testset benchmarks, particularly when using both video and ASR inputs. This demonstrates a high degree of semantic understanding.

More importantly, the capability for detailed generation does not come at the cost of performance on the fundamental chaptering task.
When comparing the segmentation metrics (temporal evaluation score F1 and tIoU) for the \textit{Short Title} task versus the more demanding \textit{Structural Info} task, we observe only a negligible difference.
For example, on VidChapter-sml300, the multimodal model achieved an F1 score of 62.4 and a tIoU of 81.6 for \textit{Short Title} generation, compared to slightly lower scores of 61.4 and 80.6 for \textit{Structural Info} generation. 
Notably, this small margin represents the largest performance gap observed across all modality inputs on both benchmarks, indicating that the model can perform complex, multi-part generation in a single forward pass without compromising its core ability to accurately segment the video.
This result strongly validates our hierarchical annotation strategy, demonstrating that training on such rich data endows the model with advanced structural reasoning capabilities.

\begin{table*}[t]
\centering
\caption{\textbf{Ablation study on the model's capability to generate hierarchical annotations.} We compare models trained with \textit{Short Title} and \textit{Structural Info} (structured chapters with short title, title, abstract, and introduction) across different input modalities (A for ASR, and V for Video) on both English (VidChapter-sml300) and Chinese (VidAtlas-testset) benchmarks. Metrics include F1 and tIoU for boundary quality evaluation, and SODA(S), CIDEr(C), as well as our proposed GRACE(G) for semantic quality evaluation.}
\label{tab:hierarchical_output_cnbench}
% Use resizebox to fit the table to the text width, useful for wide tables
\resizebox{\textwidth}{!}{%
\begin{tabular}{c|cc||ccccc||cc|ccc|ccc|ccc|ccc}
\toprule
\multirow{3}{*}{\textbf{Dataset}} & \multicolumn{2}{c||}{\multirow{2}{*}{\textbf{Modality}}} & \multicolumn{5}{c||}{\textbf{Short Title}} & \multicolumn{14}{c}{\textbf{Structural Info}} \\
\cmidrule(lr){4-22}
&  &  &  \multicolumn{2}{c|}{\textbf{Segmentation}}  & \multicolumn{3}{c||}{\textbf{Short Title}} & \multicolumn{2}{c|}{\textbf{Segmentation}} & \multicolumn{3}{c|}{\textbf{Short Title}} & \multicolumn{3}{c|}{\textbf{Title}} & \multicolumn{3}{c|}{\textbf{Abstract}} & \multicolumn{3}{c}{\textbf{Intro}} \\
\cmidrule(lr){2-3} \cmidrule(lr){4-22}
& \textbf{A} & \textbf{V} & F1 & \multicolumn{1}{c|}{tIoU} & S & C & G & F1 & \multicolumn{1}{c|}{tIoU} & S & C & \multicolumn{1}{c|}{G} & S & C & \multicolumn{1}{c|}{G} & S & C & \multicolumn{1}{c|}{G} & S & C & G \\
\midrule
\midrule

\multirow{3}{*}{\begin{tabular}{c}\textbf{VidChapter-sml300}\\ {\textbf{(English)}}\end{tabular}}  
& \cmark & \xmark & 56.5 & 78.1 & 25.9 & 148.5     &33.0& 54.8 & 77.1 & 25.5 & 147.9 &32.5&     12.8  & 91.2    & 25.6 &12.3 & 14.5 &     25.1&11.8 & 11.9&24.6  \\
  & \xmark & \cmark & 50.0 & 74.3 & 21.6 & 130.8     &27.9& 50.4 & 74.4 & 22.3 & 136.4 &28.7&      8.6 & 57.7&19.8     &8.5 &  6.4 &    19.7&8.2 & 5.2&19.4  \\
  & \cmark & \cmark & 62.4 & 81.6 & 30.1 & 190.7     &38.4& 61.4 & 80.6 & 30.8 & 194.5 &38.4&     14.6 & 107.2 &28.6    &13.4 & 14.5 &     27.4&13.0 & 10.2&27.0  \\

\midrule

\multirow{3}{*}{\begin{tabular}{c}\textbf{VidAtlas-testset}\\ {\textbf{(Chinese)}}\end{tabular}}  
& \cmark & \xmark & 58.8 & 79.7 & 24.8 & 111.3 & 28.0     & 59.1 & 79.8 & 25.5 & 112.8 & 28.6 &     16.2 & 101.7  & 27.0   & 17.5 & 57.8 &  31.8 &  16.4 & 36.0 & 29.6  \\
 & \xmark & \cmark & 57.6 & 78.9 & 21.6 & 98.1  & 25.0   & 56.8 & 78.7 & 22.0 & 97.8 & 25.1 &     12.7 & 67.4   & 21.7  & 14.5 & 37.5 & 27.3 & 13.8 & 22.2 & 25.2  \\
 & \cmark & \cmark & 66.2 & 84.0 & 30.2 & 141.5  & 34.1   & 65.9 & 83.8 & 30.8 & 143.5 & 34.6 &    18.5 & 119.8  & 30.7  & 19.1 & 66.3 &  35.3  & 18.2 & 39.8 & 33.0  \\

\bottomrule
\end{tabular}
}
\end{table*}

\subsubsection{Performance with GRPO}
\begin{table*}[t]
\centering
\caption{\textbf{Effectiveness of Reinforcement Learning with GRPO.} We compare the performance of our models before (SFT) and after applying reinforcement learning (+RL) with GRPO. The evaluation is conducted on two benchmarks across different input modalities (A: ASR, V: Video). The results show that GRPO consistently improves temporal segmentation metrics (F1, tIoU) while maintaining or slightly improving semantic quality metrics (S: SODA, C: CIDEr). Bold numbers indicate the best performance between the base model and GRPO-enhanced model for each metric.}
\label{tab:compare_grpo}
% Use resizebox to fit the table to the text width, useful for wide tables
\resizebox{\textwidth}{!}{%
\begin{tabular}{cc|cc|ccccc|ccccc}
\toprule
\multirow{2}{*}{\textbf{Method}} & \multirow{2}{*}{\textbf{Stage}} & \multicolumn{2}{c|}{\textbf{Modality}} &  \multicolumn{5}{c|}{\textbf{VidChapters7M-test}} & \multicolumn{5}{c}{\textbf{VidAtlas-test}} \\
\cmidrule(lr){3-4} \cmidrule(lr){5-9} \cmidrule(lr){10-14}
& & \textbf{A} & \textbf{V} & \multicolumn{1}{c}{F1} & \multicolumn{1}{c}{tIoU} & \multicolumn{1}{c}{S} & \multicolumn{1}{c}{C} & \multicolumn{1}{c|}{G} & \multicolumn{1}{c}{F1} & \multicolumn{1}{c}{tIoU} & \multicolumn{1}{c}{S} & \multicolumn{1}{c}{C} & \multicolumn{1}{c}{G} \\
\midrule
\midrule

\textbf{Base-asr} & \textbf{sft} & \cmark & \xmark  & {54.5} & {76.7} & \textbf{26.3} & \textbf{144.0} &  \textbf{28.9} & {58.8} & {79.7} & \textbf{24.8} & \textbf{111.3} & 28.0 \\
\textbf{GRPO-asr} & \textbf{+rl } & \cmark & \xmark  & \textbf{54.8\small{(+0.3$\uparrow$)}} & \textbf{77.2\small{(+0.5$\uparrow$)}} & 25.3\small{(-1.0$\downarrow$)} & 143.7\small{(-0.3$\downarrow$)} &  28.8 \small{(-0.1$\downarrow$)} & \textbf{59.6\small{(+0.8$\uparrow$)}} & \textbf{80.2\small{(+0.5$\uparrow$)}} & {24.7\small{(-0.1$\downarrow$)}} & {109.9\small{(-1.4$\downarrow$)}} &  28.0\small{($\uparrow\downarrow$)} \\
\midrule
\textbf{Base-vid} & \textbf{sft} & \xmark & \cmark &  {50.2} & {74.3} & {22.9} & {138.3} &  25.4 & {57.6} & {78.9} & {21.6} & {98.1} & 25.0 \\
\textbf{GRPO-vid} & \textbf{+rl } & \xmark & \cmark &  \textbf{50.6\small{(+0.4$\uparrow$)}} & \textbf{74.8\small{(+0.5$\uparrow$)}} & {22.9\small{($\uparrow\downarrow$)}} & \textbf{139.4\small{(+1.1$\uparrow$)}} &  25.4\small{($\uparrow\downarrow$)} & \textbf{58.4\small{(+0.8$\uparrow$)}} & \textbf{79.6\small{(+0.7$\uparrow$)}} & \textbf{21.9\small{(+0.3$\uparrow$)}} & \textbf{98.2\small{(+0.1$\uparrow$)}} &  25.0\small{($\uparrow\downarrow$)}  \\
\midrule
\textbf{Base-vidasr} & \textbf{sft} & \cmark & \cmark  & {59.3} & {79.6} & {30.6} & {186.6} &  34.3  & {66.2} & {84.0} & {30.2} & {141.5} & { 34.1} \\
\textbf{GRPO-vidasr} & \textbf{+rl } & \cmark & \cmark &  \textbf{60.8\small{(+1.5$\uparrow$)}} & \textbf{80.7\small{(+1.1$\uparrow$)}} & \textbf{31.0\small{(+0.4$\uparrow$)}} & \textbf{190.7\small{(+4.1$\uparrow$)}} &  \textbf{34.6\small{(+0.3$\uparrow$)}} & \textbf{66.8\small{(+0.6$\uparrow$)}} & \textbf{84.3\small{(+0.3$\uparrow$)}} & \textbf{30.4\small{(+0.2$\uparrow$)}} & \textbf{141.7\small{(+0.2$\uparrow$)}} &  \textbf{34.4\small{(+0.3$\uparrow$)}} \\
\bottomrule
\end{tabular}
}
\end{table*}
To validate the effectiveness of our GRPO-based reinforcement learning stage, we compare the performance of our models before (SFT-base) and after (+RL) this optimization. The results, detailed in Table~\ref{tab:compare_grpo}, confirm that GRPO serves as a powerful fine-tuning method for enhancing temporal precision in video chaptering.
From the experimental results, we draw three key conclusions.

First, GRPO directly and consistently improves metrics correlated with temporal segmentation accuracy. As hypothesized, by optimizing with a reward focused on temporal alignment, we observe a clear performance boost in F1 and tIoU scores across all configurations. For instance, on the VidAtlas-test set, the GRPO model with video input achieves a notable gain of +0.8 in F1 and +0.7 in tIoU over its SFT baseline. This empirically validates that GRPO effectively sharpens the model's ability to predict precise chapter boundaries.

Second, we observe a significant degree of cross-modal transferability from the RL training. Notably, despite the GRPO training being conducted exclusively on the video modality, the temporal localization performance of the ASR and Video+ASR inputs also improves. The GRPO model with Video+ASR input, for example, achieves a +1.5 F1 and +1.1 tIoU gain on VidChapter7M-test. This suggests that the optimization is not merely learning a superficial visual-to-temporal mapping but is refining a more abstract, modality-agnostic representation of temporal structure within the language model's parameters.

Finally, these enhancements in temporal precision are achieved without sacrificing semantic quality. Crucially, although our reward function is agnostic to content, semantic metric such as CIDEr remain highly comparable to the SFT baseline, and in some cases even improve (\eg, +1.1 CIDEr for video input on VidChapters7M-test.). Composite metrics like SODA and GRACE, which balance segmentation and description, also maintain their performance or exhibit slight gains. 
This indicates that the KL-regularized optimization successfully avoids policy degradation, suggesting a positive effect where more accurate segmentation enables the model to generate more focused and relevant content. 
In summary, GRPO acts as a critical fine-tuning step, effectively sharpening the model's temporal acuity while preserving its descriptive capabilities.

\subsection{Qualitative Visualization}
To provide a more intuitive understanding of our model's capabilities beyond quantitative metrics, we present qualitative examples on both English and Chinese videos.
These visualizations showcase ARC-Chapter's ability to generate accurate, coherent, and hierarchically structured outputs in multiple formats and languages.

\begin{figure}
    \centering
    \vspace{-5mm}  
    \includegraphics[width=1.0\textwidth]{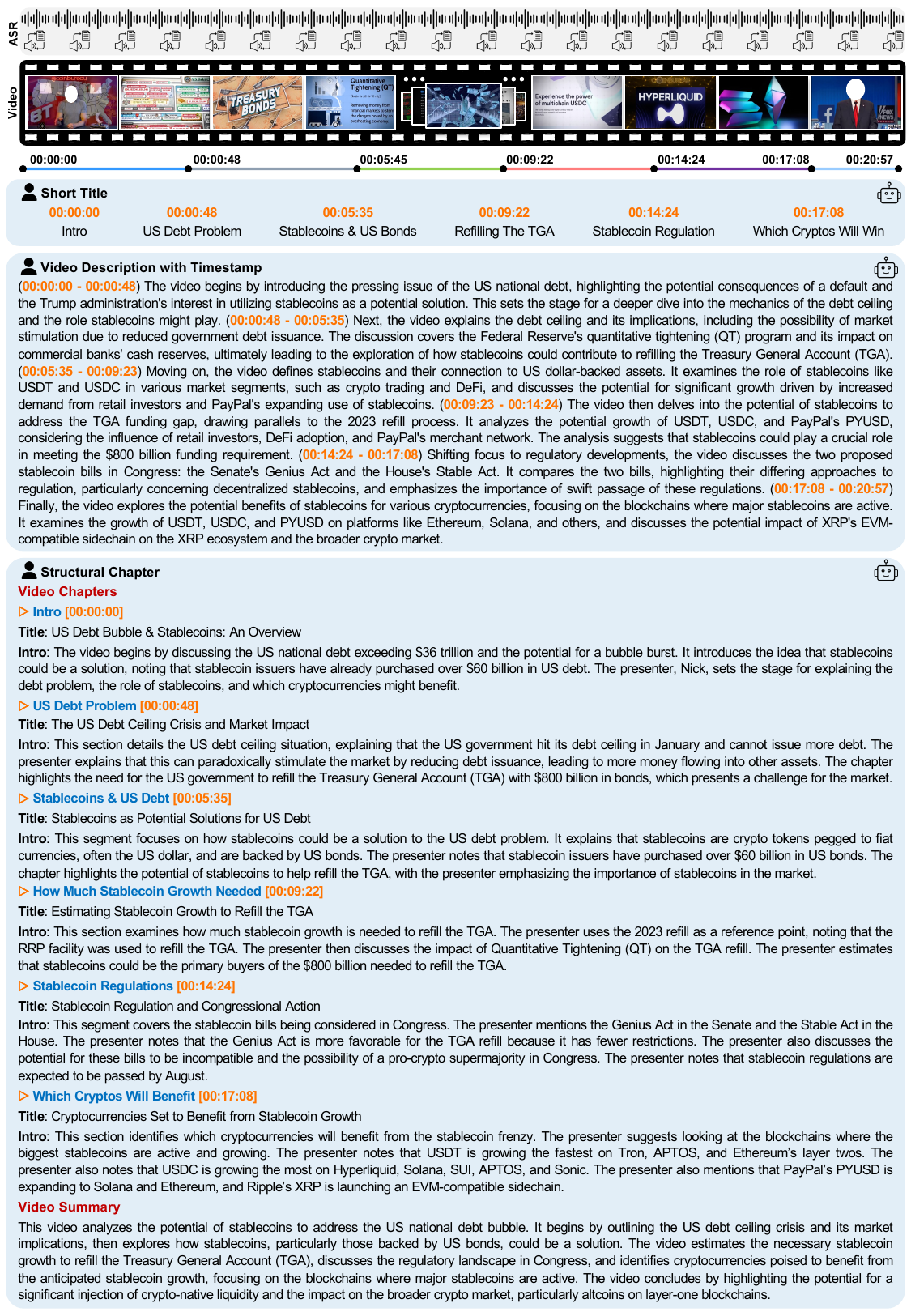}
    \vspace{-8mm}  
    \caption{Qualitative results on an English video about finance and cryptocurrency.}
  \label{fig:visulization_en} 
\end{figure}

Fig.~\ref{fig:visulization_en} illustrates the model's performance on a challenging English video discussing US debt and the role of stablecoins. 
The topic is dense with financial terminology and complex arguments. Our model successfully navigates this complexity across all output formats.
The \textit{Short Title} accurately segments the video into logical thematic units, such as "Intro", "Stablecoin Regulation".
The \textit{Video Description with Timestamp} summarizes the video content for each chapter.
More impressively, the \textit{Structural Chapters} demonstrates the model's advanced capability for hierarchical chaptering.
The generated title, abstract, and introduction for each chapter are distinct yet complementary, providing a rich, layered understanding of the content that mirrors human-authored summaries.

\begin{figure}
    \centering
    \vspace{-5mm}  
    \includegraphics[width=1.0\textwidth]{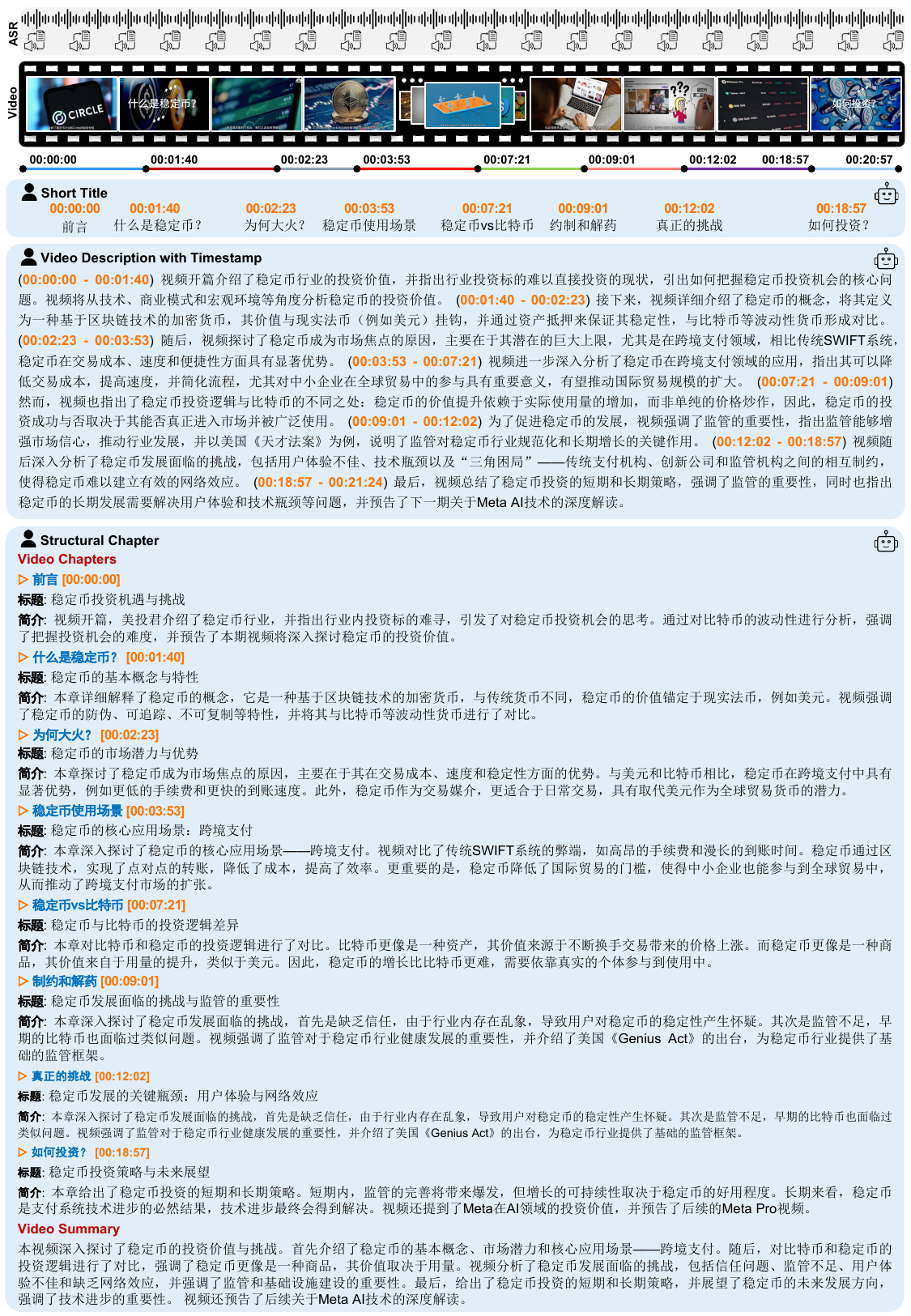}
    \vspace{-8mm}  
    \caption{Qualitative results on a Chinese video discussing stablecoins.}
  \label{fig:visulization_cn} 
\end{figure}

To showcase the multilingual performance of our model, Fig.~\ref{fig:visulization_cn} presents the results for a Chinese video on a similar topic.
The model exhibits a comparable level of understanding and generation quality in Chinese.
The generated \textit{Short Titles} are precise. The detailed \textit{Description} and \textit{Structural Chapters} are fluent and contextually appropriate.
This strong cross-lingual performance underscores the model's ability to generalize the learned chaptering and summarization skills, rather than merely memorizing patterns in a single language.

Together, these qualitative examples confirm that ARC-Chapter is not only a powerful chaptering tool but also a versatile video understanding model capable of producing rich, structured, and multilingual summaries that are both accurate and useful for end-users.

\section{Conclusion}
\label{sec:conclusion}

In this report, we introduced ARC-Chapter, a scalable and robust framework for structuring long-form videos into semantically coherent chapters and hierarchical summaries. ARC-Chapter leverages a large-scale dataset of millions of long video chapters and employs a semi-automatic annotation pipeline. These innovations advance the state of the art in video chaptering and summary generation. We also proposed the GRACE metric, which addresses the limitations of existing evaluation methods by providing a granularity-robust assessment of chapter boundaries. Experimental results show that ARC-Chapter achieves superior performance across multiple benchmarks, video durations, and languages. These findings demonstrate the framework’s effectiveness and generalizability. ARC-Chapter has strong potential to facilitate efficient content navigation, retrieval, and understanding as long-form video content continues to grow rapidly.

\clearpage
\bibliographystyle{assets/plainnat}
\bibliography{paper}

% \newpage
% \beginappendix

\end{document}